# AI-driven software for automated quantification of skeletal metastases and treatment response evaluation using Whole-Body Diffusion-Weighted MRI (WB-DWI) in Advanced Prostate Cancer

*Antonio Candito[1], Matthew D Blackledge[1], Richard Holbrey[1], Nuria Porta[1], Ana Ribeiro[2], Fabio Zugni[3], Luca D'Erme[4], Francesca Castagnoli[1,2], Alina Dragan[1,2], Ricardo Donners[5], Christina Messiou[1,2], Nina Tunariu[1,2], and Dow-Mu Koh[1,2]*

[1]The Institute of Cancer Research, London, United Kingdom

[2]The Royal Marsden NHS Foundation Trust, London, United Kingdom

[3] Department of Radiology, IEO European Institute of Oncology IRCCS, Milan, Italy

[4]Universita' del Sacro Cuore, Rome, Italy

[5]University Hospital Basel, Basel, Switzerland

Corresponding Author: Dr Antonio Candito
The Institute of Cancer Research (UK)
antonio.candito@icr.ac.uk

**Keywords:** AI-driven Imaging Analysis; Whole-Body Diffusion-Weighted MRI (WB-DWI); Quantitative Imaging Biomarkers; Metastatic Bone Disease; Treatment Response Assessment

## ABSTRACT

**Objective**: Quantitative assessment of treatment response in Advanced Prostate Cancer (APC) with bone metastases remains an unmet clinical need. Whole-Body Diffusion-Weighted MRI (WB-DWI) provides two response biomarkers: Total Diffusion Volume (TDV) and global Apparent Diffusion Coefficient (gADC). However, tracking post-treatment changes of TDV and gADC from manually delineated lesions is cumbersome and increases inter-reader variability. We developed a software to automate this process.

**Approach**: Core technologies include: (i) a weakly-supervised Residual U-Net model generating a skeleton probability map to isolate bone; (ii) a statistical framework for WB-DWI intensity normalisation, obtaining a signal-normalised b=900s/mm² (b900) image; and (iii) a shallow convolutional neural network that processes outputs from (i) and (ii) to generate a mask of suspected bone lesions, characterised by higher b900 signal intensity due to restricted water diffusion. This mask is applied to the gADC map to extract TDV and gADC statistics. We tested the tool using expert-defined metastatic bone disease delineations on 66 datasets, assessed repeatability of imaging biomarkers (N=10), and compared software-based response assessment with a *construct reference standard*, defined as multidisciplinary consensus based on ≥12 months of imaging, clinical, and laboratory follow-up (N=118).

**Main results**: Average dice score between manual and automated delineations was **0.6** for lesions within pelvis and spine, with an average surface distance of **2mm**. Relative differences for log-transformed TDV (log-TDV) and median gADC were **8.8%** and **5%**, respectively. Repeatability analysis showed coefficients of variation of **4.6%** for log-TDV and **3.5%** for median gADC, with intraclass correlation coefficients of **0.94** or higher. The software achieved **80.5%** accuracy, **84.3%** sensitivity, and **85.7%** specificity in assessing response to treatment. Average computation time was **90s** per scan.

**Significance**: Our software enables reproducible TDV and gADC quantification from WB-DWI scans for monitoring metastatic bone disease response, thus providing potentially useful measurements for clinical decision-making in APC patients.

## 1. Introduction

Bone is the most common site for cancer spread in patients with Advanced Prostate Cancer (APC), with 90% of these patients developing metastatic bone disease. In up to 45% of these cases, the skeleton is the sole site of metastases [1], [2]. Metastatic bone disease significantly increases morbidity, leading to severe complications such as bone fractures and spinal cord compression, and is associated with a low five-year survival rate [3]. Although innovative targeted therapies have shown promise in clinical trials by improving quality of life, preventing complications, and extending survival for APC patients [4]–[6], treatments ultimately stop being effective. Therefore, it is imperative for clinicians to use advanced diagnostic tools to determine when to switch to a different drug to maintain the therapeutic effect, with the aim of keeping the patients "better for longer".

Conventional medical imaging techniques, such as radionuclide bone scan (BS) and computed tomography (CT), are widely used to assess metastatic bone disease [7], [8]. However, these techniques have low sensitivity in detecting malignant bone lesions, lack response criteria and are not reliable in identifying early signs of disease progression [9]–[11]. Prostate specific membrane antigen positron emission tomography (PSMA-PET) offers higher sensitivity for detecting metastatic bone lesions and evaluating disease extent in APC patients [12], [13]. Quantitative imaging biomarkers such as maximum standardised uptake value (SUV_max) and total tumour volume (TTV) can be derived from PSMA-PET by segmenting lesions with uptake above a predefined SUV threshold [14]. These biomarkers allow quantitative monitoring of disease burden after treatment; however, they remain indirect measures, reflecting radioactive tracer uptake. Moreover, PSMA-PET is costly and prone to false positives due to the treatment-related flare phenomenon [15].

In contrast, Whole-body Diffusion-Weighted MRI (WB-DWI) provides quantitative imaging biomarkers which directly reflect changes in bone cellularity [16]. WB-DWI is a non-invasive, radiation-free technique that provides excellent contrast between normal bone and malignant bone disease, enabling "at a glance" visual assessment of cancer spread throughout the skeleton with higher sensitivity than BS and CT, including detecting early disease progression [17]. On DWI, malignant bone disease appears high signal intensity on high b-values (typically b=900 s/mm², b900) images, due to restricted water diffusion within cancerous tissue, associated with increased cell density [18]. In addition, DWI allows for the quantification of two response biomarkers: the Total Diffusion Volume (TDV, in millilitres), reflecting tumour burden, and the global Apparent Diffusion Coefficient (gADC), calculated from two or three b-value images, which inversely correlates with tumour cellularity [19], [20]. Quantitative TDV and gADC have been shown to corroborate with response by laboratory test and clinical findings, providing objective measurements of treatment response in APC patients [21], [22]. International guidelines, including Prostate Cancer Working Group 3 (PCWG3) [7] provides recommendations for soft tissue and lymph-node disease evaluation, while METastasis Reporting and Data System for Prostate Cancer (MET-RADS-P) [23] extends these criteria to whole-body MRI for evaluating response to treatment in patients with skeletal metastases.

However, a factor limiting the use of quantitative WB-DWI measurements is the lack of advanced software that enables automated analysis and rapid calculation of metastatic disease throughout the body. Manual delineation for TDV and gADC measurements is tedious, often taking an hour or more depending on the disease volume, making it impractical for clinical application [24]. The NCI's Quantitative Imaging Network (QIN) [25] is already promoting the development and clinical validation of quantitative imaging tools aimed at measuring and predicting tumour response.

Semi-automated tools have been developed to facilitate the delineation and summation of metastatic bone disease [26]–[29]. However, multiple manual steps are still required, including selecting WB-DWI sequences as input from the imaging study, adjusting contrast on b900 images to differentiate/delineate suspected bone lesions from background bone marrow, and refining the generated mask to remove non-skeletal structures with similar signal intensity than bone lesions (e.g., adjacent internal organs, spinal canal contents, nerve roots, lymph nodes). These manual steps can incur delays, disrupt radiologists' reporting workflow, and increase the inter-reader variability of measuring the TDV and gADC.

In this study, we developed and test a signal-based AI-driven software solution to automatically identify, segment and quantify metastatic bone disease in patients with APC. We hypothesise that automated delineation of bone regions with restricted diffusion could be used to estimate TDV and the associated gADC from pre- and post-treatment WB-DWI scans, thus be used to assess disease response of metastatic bone disease. To test this hypothesis, we evaluated the accuracy of our software using retrospective and prospective WB-DWI scans in APC patients for assessing treatment response. Furthermore, we derived and tested a *Response Evaluation Criteria* (REC) based on our automated TDV and gADC measurements and their percentage changes post-treatment. Our quantitative measurements could assist clinicians by providing a more objective paradigm for classifying patients as responders, stable, or with disease progression, thereby improving clinical decision-making.

## 2. Methods

### 2.1 Patient population

This study was approved by research and ethics committee (Committee for Clinical Review (CCR) - Royal Marsden NHS Foundation Trust), which waived the requirement for patient consent for both retrospective and prospective datasets. All data were fully anonymised, and the study was conducted in accordance with the Declaration of Helsinki (2013). The inclusion criteria included patients with confirmed APC with bone metastases who were about to start systemic treatment or change to a new line of therapy, with no contraindications for whole-body MRI. Additionally, patients with metal implants (e.g., hip replacements) or other source of imaging artefact (e.g., anterior thoracic signal loss, ghosting, and geometric distortion) were not excluded from the study.

**Table 1** summarises patient characteristics in the datasets used to develop the signal-based AI-driven software solution. The training set consisted of a single-centre retrospective cohort of 43 patients (**Dataset A**) who underwent pre- and post-treatment WB-DWI. A certified radiologist (5+ years in functional cancer imaging) delineated all metastatic bone disease on the b900 DWI sequence, using b900 images and gADC maps, corroborated with T1-weighted (T1w) Dixon images [30]. To assess the repeatability of automated TDV and gADC measurements, we used a cohort of 10 patients (**Dataset B**) who underwent two baseline scans on the same day, using the same scanner and protocol, prior to treatment initiation. For testing, we employed a retrospective multi-centre cohort (**Datasets C**) and a single-centre prospective cohort (**Dataset D**, acquired from the same institution as **Dataset A**) consisting of 102 and 16 patients, respectively, all of whom underwent pre- and post-treatment WB-DWI. In **Dataset C,** a consultant radiologist (15+ years of experience in whole-body MRI and clinical trials) delineated all metastatic bone lesions on the b900 DWI sequences, using the same multiparametric approach as in **Dataset A**, for a subset of 33 patients (66 scans). Across all datasets, patients underwent follow-up imaging approximately 12 weeks after initiating systemic therapy.

| | **Pre-treatment** | **Post-treatment** | **p-value** |
|---|---|---|---|
| ***Training*** | | | |
| **Dataset (A)** | | | |
| Retrospective single-center cohort<br>43 patients / 86 WB-DWI scans | | | |
| **WB-DWI response biomarkers**[*]<br>**(manual delineations)** | | | |
| TDV [mL] | 164 [32 - 813] | 155 [32 - 927] | 0.02 |
| gADC $[\text{x}10^{-3}\ mm^2/s]$ | 0.864 [0.66 - 1.147] | 0.917 [0.68 - 1.268 | 0.001 |
| | | | |
| ***Test*** | | | |
| **Dataset (B)** | | | |
| Retrospective single-center cohort<br>10 patients / 20 WB-DWI scans<br>(Repeatability study) | | | |
| | | | |
| **Dataset (C)** | | | |
| Retrospective multi-center cohort<br>102 patients / 204 WB-DWI scans | | | |
| **WB-DWI response biomarkers**[*]<br>**(manual delineations - 66 scans)** | | | |
| TDV [mL] | 78 [50 - 285] | 92 [61 - 325] | <0.001 |
| gADC $[\text{x}10^{-3}\ mm^2/s]$ | 0.926 [0.77 - 1.127] | 1.006 [0.808 - 1.267] | <0.001 |
| **Response Assessment**[†]<br>**(constructed reference standard)** | | | |
| Benefit (responders/stable) | | 59/102 (57.8%) | |
| No Benefit (progression) | | 43/102 (42.2%) | |
| | | | |
| **Dataset (D)** | | | |
| Prospective single-center cohort<br>16 patients / 32 WB-DWI scans) | | | |
| **Response Assessment**[†]<br>**(constructed reference standard)** | | | |
| Benefit (responders/stable) | | 11/16 (68.6%) | |
| No Benefit (progression) | | 5/16 (31.4%) | |

**Table 1.** Training and test cohorts used for software development. WB-DWI response biomarkers were derived from expert manual delineations of metastatic bone disease. Response assessment was based on a *construct reference standard* based on clinical assessment, laboratory findings and, all available imaging up to one year after treatment. Pre- and post-treatment TDV and gADC values were compared using the Wilcoxon signed-rank test, with a p-value <0.05 considered statistically significant. [†]Data are numbers of patients, with percentages in parentheses. [*]Data are medians, with 1st and 3rd quartiles in parentheses.

### 2.2 *Reference standard* for response to treatment

A consensus panel of experienced radiologists and oncologists were used to determine the *construct reference standard* for assessing response to treatment, reviewing all available diagnostic imaging, laboratory test results, and clinical findings up to one year from treatment initiation (see **Table 1**). In **Datasets C** and **D** APC patients were classified as *Benefit* (responders/stable) or *No Benefit* (progression) from treatment. This dichotomy was designed to evaluate the software for: (i) potential use in clinical trials assessing the therapeutic effects of novel targeted therapies [31]; and (ii) guiding therapy discontinuation in cases of early signs of disease progression.

### 2.3 Imaging protocol

We used a well-accepted MET-RADS-P compatible [23] whole-body MRI protocol for staging and response assessment of metastatic bone disease including morphological and DWI sequences: sagittal T1/T2 spine, 3D T1w DIXON, T2 HASTE, and DWI with either two (50/900 s/mm²) or three (50/600/900 s/mm²) b-value images [32]. In our study, images were acquired between 2015 and 2023 using a 1.5T scanner (MAGNETOM Aera/Avanto, Siemens Healthcare, Erlangen, Germany). WB-DWI sequences covered 5 or 6 stations, extending either from the skull base to mid-thigh or from the skull vertex to the knees. Each station consisted of 40 slices with a slice thickness of 5mm. WB-DWI was performed using axial 2D single-shot echo-planar imaging with acceleration factor 2 (GRAPPA) and three/four scan trace encoding directions. gADC maps were calculated by fitting a mono-exponential decay model to the b-value images [33], [34]. All MRI parameters related to the acquisition of WB-DWI sequences for each dataset involved in the study are reported in the Supplementary Material (Tables and Figures, **Table S1**).

### 2.4 Signal-based AI-driven software

We implemented a two-step AI-driven approach: first, a weakly-supervised deep-learning model localises the skeleton within the WB-DWI scan; second, a supervised shallow convolutional neural network (CNN) delineates suspected bone lesions within this region, enabling automated quantification of metastatic bone disease. To achieve full automation, these AI models are integrated with pre-processing methods for automatic DWI series selection and signal harmonisation, and post-processing algorithms to minimise false positives, followed by extraction of imaging biomarker measurements from each WB-MRI study.

The software reads paired pre- and post-treatment whole-body MRI studies, automatically identifying and processing WB-DWI sequences for each timepoint separately, and generates a structured report as output (see exemplar in Supplementary Material, **Figure S1**). This report includes automated delineations of suspected bone lesions with restricted diffusion properties on DWI (including areas of possible T2 shine through [16]), along with TDV and gADC measurements and their absolute and percentage changes after treatment. **Figure 1** shows the main components of the workflow:

Step 1 - Workflow composition (Pre-processing)

Automatically selects DWI series from the WB-MRI study using DICOM metadata, aligns and resamples multi-b-value series, and calculates b=0 s/mm² (S0) image and gADC map.

Step 2 - Anatomical localisation

A previously validated weakly-supervised 3D patch-based Residual U-Net, trained on 532 WB-DWI scans with atlas-derived soft-label segmentations, outputs multi-class voxel-wise probability maps of the skeleton, adjacent internal organs (bladder, kidneys, liver and spleen), and spinal canal [35], [36]. The model employs a symmetric encoder–decoder architecture with five convolutional layers, starting with 32 filters, and was trained using randomly sampled patches of size 128x128x64 voxels from the full volume. It takes two input channels, the calculated S0 image and gADC map. The body region probability maps are used to isolate bony regions from adjacent soft tissue organs and image background. The spinal canal delineation is then used to enable signal harmonisation (Step 3).

Step 3 - Signal harmonisation

Signal harmonisation is performed through histogram-based normalisation of voxel intensities from the high b-value sequence within the spinal canal mask from Step 2, generating a signal-normalised b900 image that improves inter- and intra-patient variability in signal intensity and enhances contrast between suspected hyperintense bone lesions and normal bone marrow [37].

Step 4 – Bone lesion delineation

A supervised shallow CNN trained on 86 WB-DWI scans (**Dataset A**) with two input channels from Steps 2-3, (i) the skeleton map and (ii) the signal-normalised b900 image, is used to generate a bone delineation mask identifying areas indicative of restricted diffusion. Input volumes were interpolated to 256x256 at 1.6mm in-plane resolution, and 64x64x64 patches were randomly sampled for training. The network consists of three convolutional layers (16, 32, 64 filters), each with ReLU activation, batch normalisation, and 0.2 dropout; the final output layer uses sigmoid activation. The Adam optimiser (learning rate $10^{-3}$) was employed to minimise a weighted combination of dice and cross-entropy losses.

Step 5 - Post-processing

The mask is refined to address potential misclassifications from the CNN by removing regions of interest (ROIs) with >65% of voxels showing gADC <0.5 $\times 10^{-3}$ mm$^2$/s (to filter out signals from normal fatty bone marrow or noise) or overlapping with non-skeletal organs.

Step 6 - Feature extraction

The resulting mask is then applied to the gADC map to extract TDV and gADC statistics for global and individual skeletal regions, including limbs, pelvis, thorax, and lumbar, thoracic, and cervical spine.

Finally, extracted measurements are analysed to assess treatment response using cutoff values for TDV and median gADC percentage changes, with optimal thresholds determined through bootstrapping and grid search to maximise Youden's index (detailed in Sections 2.5.4 and 2.6.4).

The full implementation details for the software pipeline and AI models are available in the Supplementary Material, "Technical Documentation".

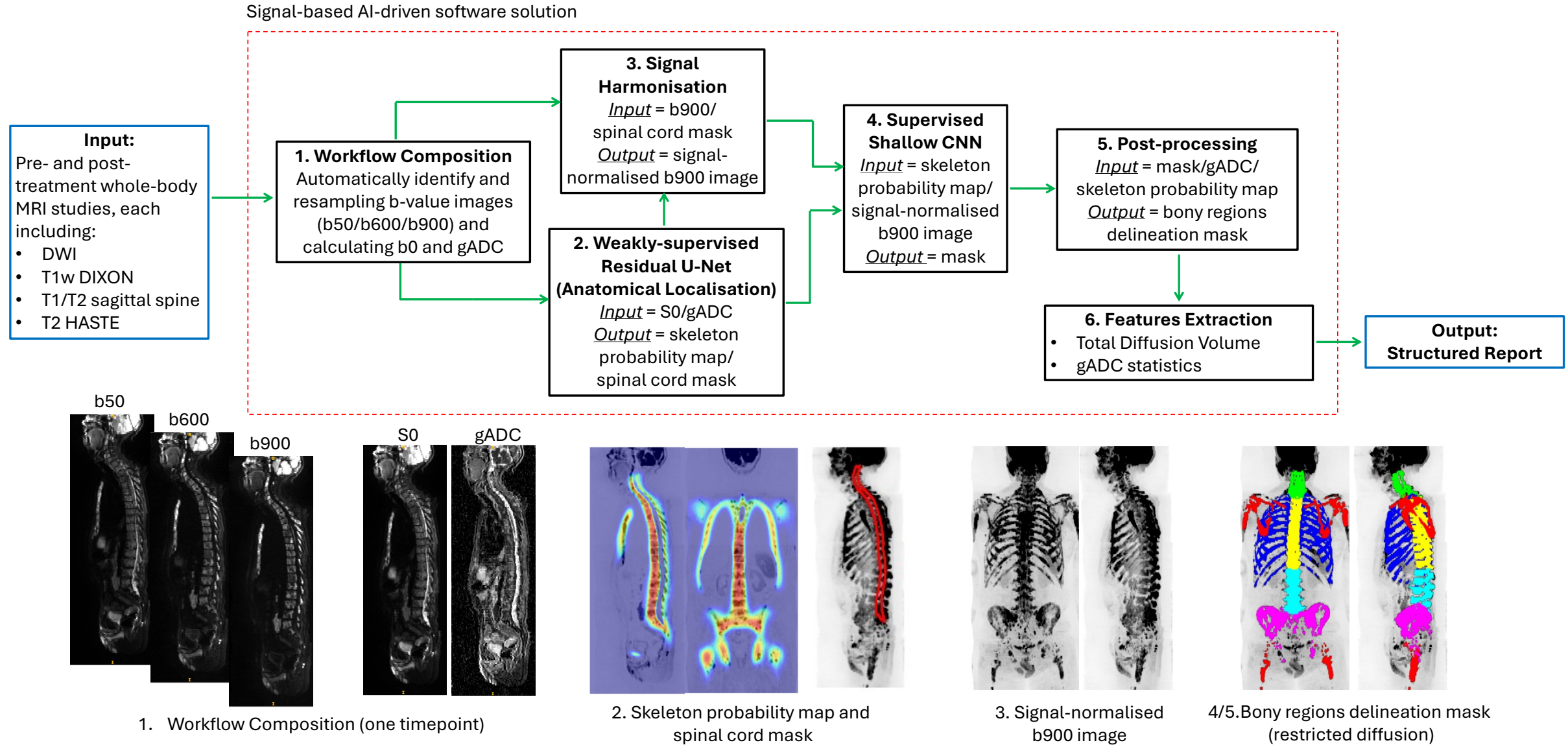

**Figure 1.** Flow diagram of the proposed signal-based AI-driven software solution. The software processes pre- and post-treatment whole-body MRI studies to generate a structured report with automated delineation of bony regions characterised by restricted diffusion (high b900 signal intensity and low ADC values), along with associated TDV and gADC measurements. In the example shown, the patient presents with diffuse metastatic bone disease involving most of the skeleton, which is correctly delineated by the tool. For each step in the diagram, input and output images are shown alongside the methods or algorithms used to process them. The pipeline includes: (1) automatic DWI series selection and pre-processing; (2) anatomical skeleton localisation through a weakly-supervised Residual U-Net; (3) DWI signal harmonisation based on spinal canal delineation; (4) delineation of hyperintense bone lesions using a supervised shallow CNN; (5) post-processing to remove potential false positives, such as ROIs with ADC values that may reflect fatty bone marrow, and those located outside skeletal regions; and (6) feature extraction for biomarker quantification. The entire process requires approximately three minutes for both timepoints.

### 2.5 Software development and testing

#### 2.5.1 Supervised shallow CNN training

We aimed to develop an AI-based model to delineate bone lesions showing restricted diffusion, with the following characteristics: (i) reduced model complexity for efficient training and deployment of a lightweight network, (ii) fast inference on CPU-based workstations without the need for GPU hardware, (iii) no requirement for large annotated datasets, and (iv) retained interpretability by delineating ROIs with a consistent pattern across scans, high b900 signal intensity and low ADC within the patient skeleton. To this end, we implemented a supervised shallow CNN (as described in Section 2.4, Step 4), instead of a deeper model (e.g., nnU-Net). The CNN also delivered smoother, more focused boundaries, reducing noise and artefacts for higher-quality segmentations compared with threshold-based methods, which typically generate coarse masks.

We performed 5-fold cross-validation to train the shallow CNN using all WB-DWI scans in **Dataset A**, ensuring that images from the same patient were included in the same training/validation set. For each fold, we calculated the dice score, precision, recall, and average surface distance between automated and expert-defined metastatic bone disease delineations. Additionally, we assessed the relative differences in TDV and median gADC measured from manual and automated methods. TDV was log-transformed (log-TDV) to reduce the scaling effects of errors caused by variability in disease volumes within the patient cohort. The final model was selected based on the cross-validation fold that achieved the highest dice score and the lowest relative error for log-TDV and median gADC.

#### 2.5.2 Accuracy of automated WB-DWI delineation on test datasets

We tested the signal-based AI-driven software solution using 66 WB-DWI scans in **Dataset C**. Automated delineation accuracy was evaluated (i) considering all disease delineations across the skeleton, (ii) for individual skeletal regions, including the limbs, pelvis, lumbar/thoracic/cervical spine, and ribcage, (iii) along with its correlation with manually measured TDV. We reported the same overlap-based metrics and WB-DWI parameters as defined in the cross-validation approach.

#### 2.5.3 Repeatability of automated WB-DWI response biomarkers

Using our software, we generated delineations for all 20 baseline scans in the repeatability cohort (**Dataset B**) and applied them to the corresponding gADC maps to estimate TDV and gADC statistics. We assessed the repeatability of log-TDV as well as mean, median, variance, skewness, and kurtosis for gADC. For all parameters, we reported the coefficient of variation (CoV), repeatability coefficient (RC), within- and between-subject standard deviations (Sw and Sb), and the intraclass correlation coefficient (ICC) [38]. Bland-Altman plots [39] were generated for all parameters and analysed to identify any systematic biases and random errors in measurements of TDV and gADC. Quantitative imaging biomarkers with CoV and RC values below 5% and 15%, respectively, and ICC values above 0.9 will be considered for further investigation in assessing response to treatment.

#### 2.5.4 Accuracy in classifying response to treatment on test datasets

**Table 2** shows the proposed REC for quantitatively assessing response to treatment for APC patients based on percentage changes after treatment in TDV (ΔTDV) and median gADC (Δmedian gADC):

1. *Responders*: a significant decrease in ΔTDV (≤ –40%) [40] and/or a significant increase in Δmedian gADC (≥ +25%) [41], consistent with expected clinical response patterns.
2. *Progression*: If the above criteria are unmet, a significant increase in ΔTDV (> +40%) [26] or an increase in delineated regions compared to the baseline scan.
3. *Stable*: If neither condition is met.

Using our software, we generated automated delineations and extracted TDV and gADC measurements from pre- and post-treatment WB-DWI scans for all patients in **Datasets C** and **D**. The proposed response criteria were applied to these datasets to: (i) test and validate the pre-defined ΔTDV and Δmedian gADC cutoff values, and (ii) assess accuracy, sensitivity, and specificity by comparing software-based response classification (responder, stable, or progressing) with the *construct reference standard* [42] (Section 2.2).

Post-treatment percentage change in TDV

Post-treatment percentage change in median gADC

| | **Significant Increase**<br>**> +40% or**<br>**≥ +10 ROIs (> 1mL) or**<br>**≥ +6 ROIs (> 3mL)** | **No change** | **Significant Decrease**<br>**≤ −40%** |
|---|---|---|---|
| **Significant Increase**<br>**≥ +25%** | Check auto delineations or date of baseline MRI in relation to treatment initiation | ***<u>Benefit - Responder</u>***<br>T2-shine through | ***<u>Benefit - Responder</u>***<br>tumor cell death results in initial increased of water diffusivity |
| **No change** | ***<u>No Benefit - Progression</u>***<br>An increase in the geographic spread of the same type of tumor, along with a higher density of tumor cells | ***<u>Benefit - Stable</u>***<br>Slight alteration in tumor cellularity,<br>minor progression and/or minor response | ***<u>Benefit - Responder</u>***<br>ADC measured from normal bone marrow / areas of scarring,<br>minor areas of non-responding disease |
| **Decrease** | ***<u>No Benefit - Progression</u>***<br>An increase in the geographic spread of the same type of tumor, along with a higher density of tumor cells | ***<u>Benefit - Stable</u>***<br>Slight alteration in tumor cellularity,<br>minor progression and/or minor response | ***<u>Benefit - Responder</u>***<br>ADC measured from normal bone marrow / areas of scarring,<br>minor areas of non-responding disease |

**Table 2.** *Response Evaluation Criteria* (REC) for classifying APC patients as either *Benefit* (responders/stable disease) or *No Benefit* (disease progression) based on automated measurements of TDV and median gADC. This approach accounts for possible combinations of TDV and median gADC changes at follow-up. These changes are based on whole-body DWI sequences (b-value images and gADC maps); and recommended to be used alongside the qualitative assessment by radiologists.

### 2.6 Statistical Analysis

Values are reported as medians with interquartile ranges (IQRs) for continuous variables. A p-value<0.05 was considered statistically significant. All analyses were performed using Python 3.9 with Scikit-learn 1.5.0, Statsmodels 0.14.2, and Scipy 1.13.1.

#### 2.6.1 TDV and gADC comparisons

TDV and gADC distributions before and after treatment, based on expert-defined metastatic bone disease delineations, were compared using the Wilcoxon signed-rank test.

#### 2.6.2 Accuracy of automated WB-DWI delineation on test datasets

The correlation between overlap-based metrics (dice score, precision and average surface distance) and manually measured TDV was assessed using the Spearman correlation coefficient (*r*), interpreted as follows: *r*=0.4-0.6 (moderate), *r*=0.6-0.8 (strong), and *r*≥0.8 (very strong).

#### 2.6.3 Repeatability of automated WB-DWI response biomarkers

Repeatability of log-TDV and gADC statistics was evaluated according to QIBA recommendations [43]. A paired two-tailed t-test assessed differences between first and second baseline measurements, with 95% confidence intervals (CIs) calculated for all parameters [38]. If no significant difference was found, a bias of zero (bias=0) was assumed, with the RC equal to the 95% limits of agreement (LoA). ICC values were interpreted as ICC<0.5 (poor), 0.5-0.75 (moderate), 0.75-0.9 (good), and >0.9 (excellent).

#### 2.6.4 Accuracy in classifying response to treatment on test datasets

Bootstrapping was used for two purposes: (i) to identify optimal ΔTDV and Δmedian gADC cutoff values, and (ii) to evaluate the classification performance difference between these fitted thresholds and the pre-defined REC-based thresholds (Section 2.5.4). At each iteration, cutoff pairs were evaluated via a grid search, with binary classifications compared to the *construct reference standard* to identify the combination resulting in the highest Youden's index (above 0.6 indicative of acceptable performance). 95% CIs for the optimal cutoff values, Youden's index, and area under the curve (AUC) values were derived from 200 bootstrap iterations on resampled test datasets.

Differences in pre- and post-treatment automated measurements of TDV and median gADC for patients classified as responders or with disease progression were assessed using the Wilcoxon signed-rank test (two-tailed for TDV, one-tailed for median gADC).

Finally, 95% CIs for accuracy, sensitivity, and specificity were calculated using Wilson score interval to assess agreement between the *construct reference standard* and software-based response classification. A diagnostic accuracy of at least 80% was defined as the primary endpoint for evaluating the successful implementation of our software solution.

## 3. Results

### 3.1 Patient characteristics

gADC values from expert-defined metastatic bone disease delineations align with literature-reported values for APC patients [44], [45] (**Table 1**). In **Dataset A**, TDV measured 164mL [32-813] pre-treatment and 155mL [32-927] post-treatment, with gADC ranging from 0.66 to 1.147x$10^{-3}$mm$^2$/s and 0.68 to 1.269x$10^{-3}$mm$^2$/s, respectively. In **Dataset C**, TDV was 78mL [50–285] pre-treatment and 92mL [61-325] post-treatment, with consistent gADC distributions to **Dataset A**. The higher TDV values in **Dataset A** reflect our intentional inclusion of patients with a significant extent of metastatic bone disease for model training, ensuring that the shallow CNN was exposed to sufficient positive examples to learn patterns for delineating suspected bone lesions from WB-DWI scans.

### 3.2 Accuracy of automated WB-DWI delineation on test datasets

The computation time for executing the software pipeline per whole-body MRI scan was 90 seconds on a 2.4 GHz Quad-Core Intel Core i5 processor. **Figure 2** presents the overlap-based metrics between manual and automated delineations, along with the relative error of log-TDV and median gADC across 66 WB-DWI scans (**Dataset C)**. A patient-wise dice score of 0.52 [0.34-0.65] was obtained, with most of the error originating from the thorax region, where the score was lower than 0.35. For the upper and lower limbs, the dice score was 0.60 [0.44-0.69] with an average surface distance of 2.96mm [0.75-7.5]. In the pelvis, 0.58 [0.36-0.73] and 3.94mm [0.88-11.8], respectively. For the spine, 0.60 [0.43-0.74] and 1.55mm [0.87-4.55], respectively. The relative differences in log-TDV and median gADC between manual and automated delineations were 8.82% [4.79-22.6] and 4.97% [1.39-16.9], respectively. The highest values for the relative difference in TDV were observed in the thorax, with some variability noted in the lower limbs and cervical spine compared to other skeletal regions. **Figure 3** provides exemplar gADC maps and b900 images from four patients in **Dataset C**, showing both manual and automated delineations of suspected bone lesions.

Nonetheless, we observed moderate-to-strong correlations between automated delineation accuracy and manually measured TDV (**Figure 4**). Increasing TDV correlated positively with dice score (r=0.55, p<0.001) and precision (r=0.62, p<0.001), and negatively with average surface distance (r=–0.70, p<0.001) and relative median gADC difference (r=–0.54, p<0.001).

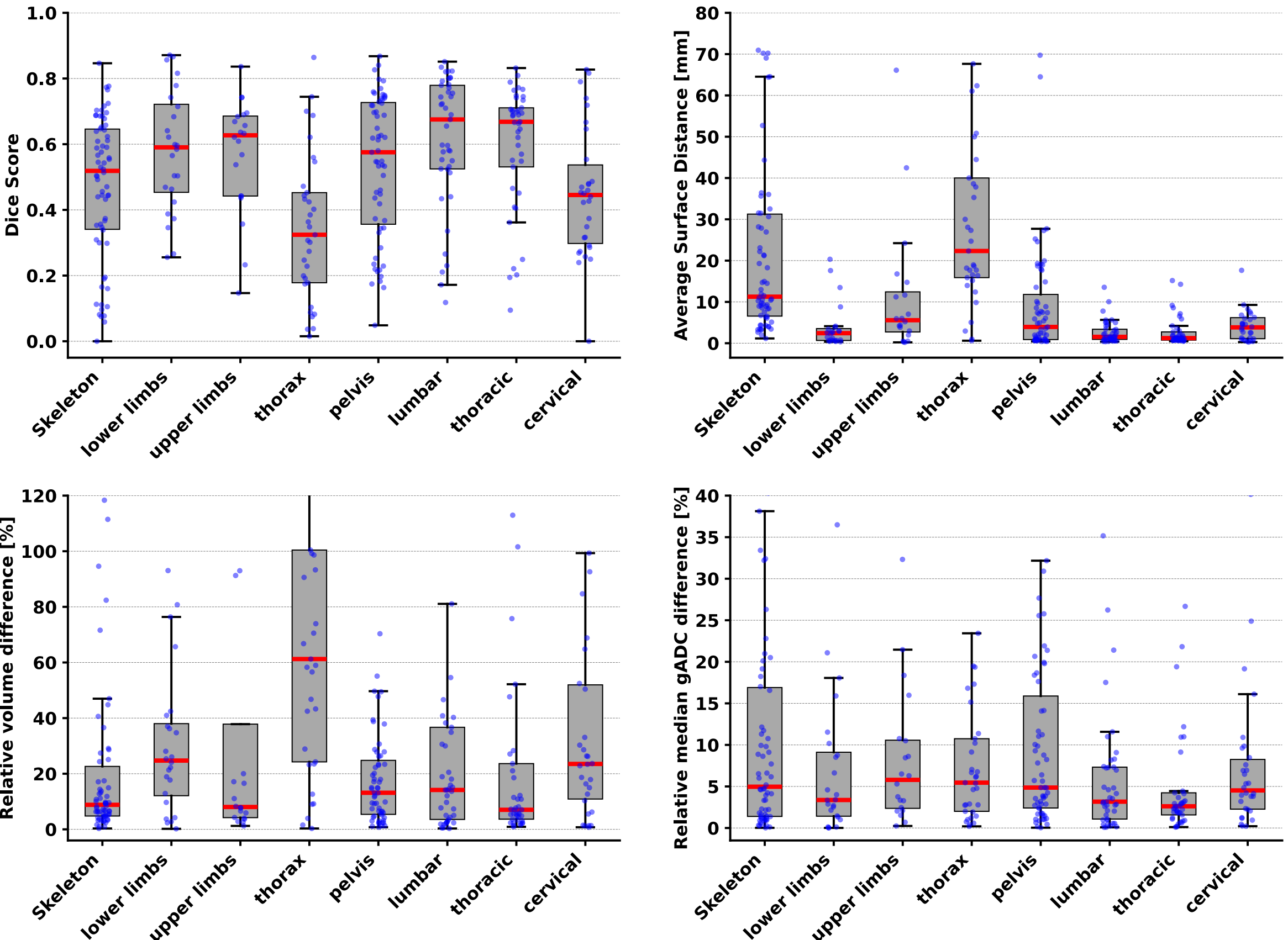

**Figure 2.** Software accuracy in delineating suspected bone lesions across 66 WB-DWI test datasets (**Dataset C**). Overlap-based metrics (dice score and average surface distance), along with relative differences of volume (log-transformed) and median gADC between automated and expert-defined manual delineations of metastatic bone disease, were assessed for both whole-skeleton delineations and individual skeletal regions (lower/upper limbs, thorax, pelvis, and lumbar/thoracic/cervical spine). The number of datapoints varied between boxplots, reflecting skeletal regions without metastases; in those cases, identified by the absence of manual annotations, metrics were not computed. Good agreement was observed for lesions detected in the limbs, pelvis, and spine (average dice score of 0.6), while performance was lower in the ribcage (dice score below 0.35).

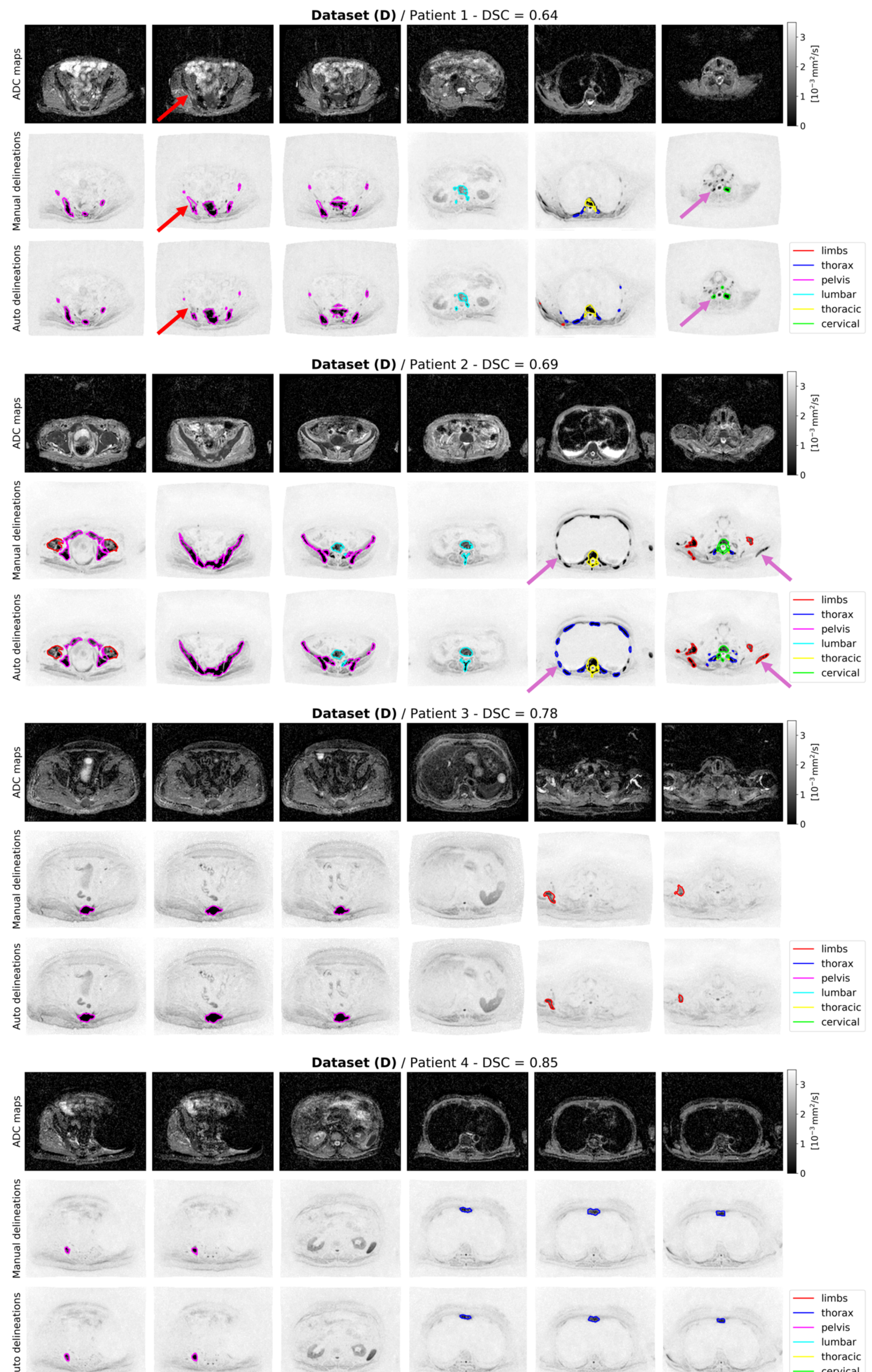

**Figure 3**. Axial gADC maps and b900 images with superimposed automated and expert-defined manual delineations of metastatic bone lesions from four patients in **Dataset C**. The software detected/delineated suspected bone lesions with high signal intensity on b900 images and low gADC values (typically above 0.5x10$^{-3}$mm$^2$/s), while excluding adjacent internal organs and image background. Overall, our solution demonstrated good agreement with the manual method, as indicated by a dice score above 0.64 for these cases. However, false negatives (red arrows) were observed in ***Patient 1***, where the manually delineated lesion in the pelvis showed low signal on the b900 image and high gADC values (above 1.4x10$^{-3}$mm$^2$/s, likely a treated lesion), suggesting potential inter-reader variability in defining active bone disease. Additionally, the automated mask showed minor imperfections by incorrectly delineating cervical spine nerves (purple arrows). For ***Patient 2***, false positives were observed in the ribs and shoulders due to increased normal bone marrow DWI signal (purple arrows). ***Patient 3*** presented only two metastatic bone lesions, one in the coccyx and one at the top of the shoulders, with suppressed DWI signal from normal bone marrow. Nevertheless, the software-generated delineations showed excellent agreement with the manual method (dice score of 0.78). For ***Patient 4***, with a manually measured TDV of 75mL, excellent agreement was observed between manual and automated methods (dice score of 0.85).

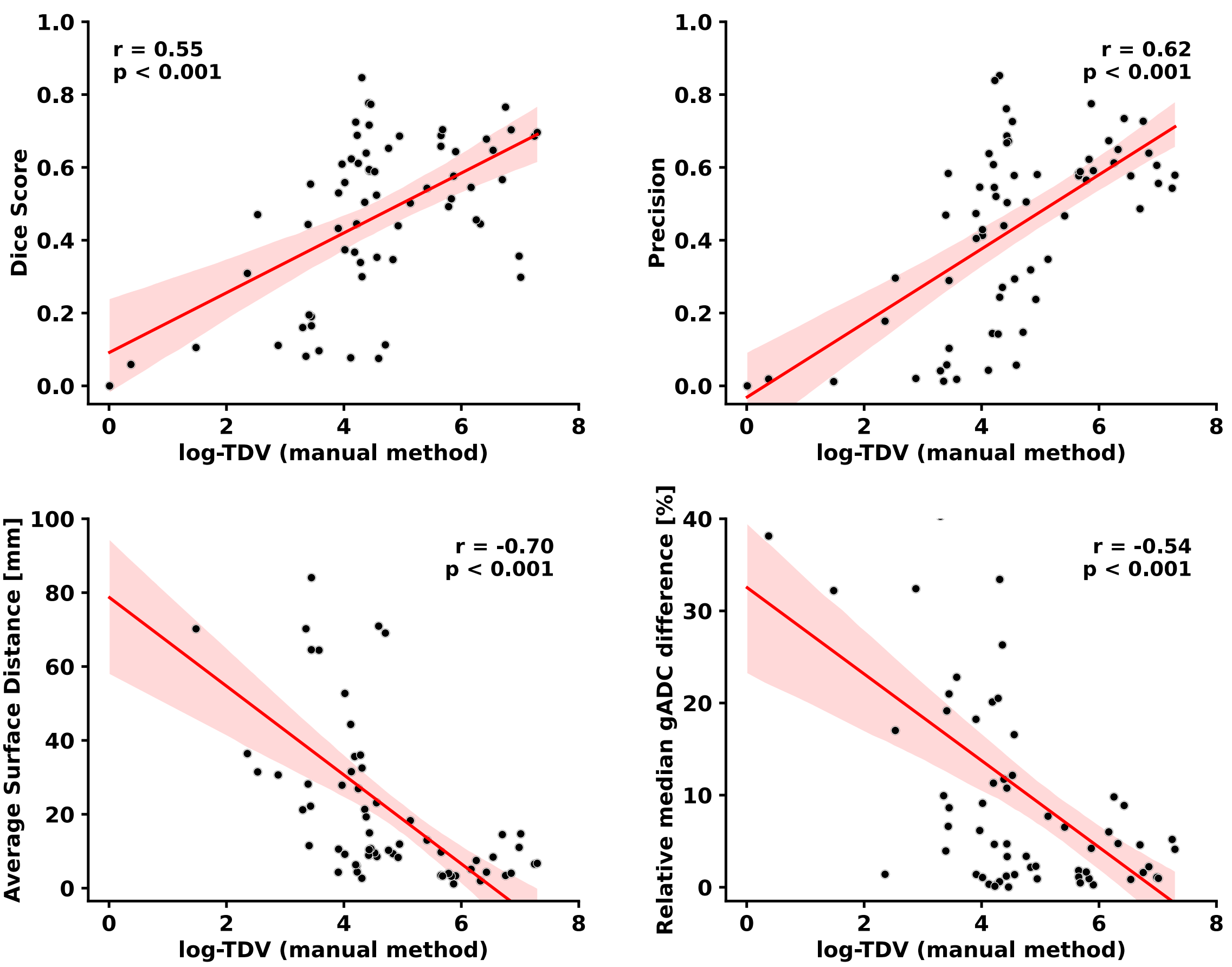

**Figure 4.** Correlation between software accuracy in delineating suspected bone lesions and manually measured TDV across 66 WB-DWI test datasets (**Dataset C**). Correlations were assessed using Spearman's correlation coefficient (*r*), with $p<0.05$ considered statistically significant. Accuracy improved with increasing disease burden. The red line represents the linear fit across all data points, while the shaded red region indicates the standard deviation. For instance, patients with TDV above 55 mL (log-TDV=4) demonstrated a relative median gADC difference below 12%.

### 3.3 Repeatability of automated WB-DWI response biomarkers

Among all tested parameters, log-TDV and median gADC demonstrated the best repeatability (see **Table 3**). The CoV for log-TDV and median gADC was 4.57% [95% CIs: 3.19-8.01] and 3.54% [95% CIs: 2.47-6.21], respectively. The RC was 12.66% [95% CIs: 8.84-22.2] for log-TDV and 9.81% [95% CIs: 6.86-17.2] for median gADC. The ICC for both parameters exceeded 0.9, with 95% CIs of 0.81-0.99 for log-TDV and 0.9-0.99 for median gADC. Bland-Altman plots (**Figure S2**) showed no consistent trend suggesting systematic errors in the measurements. The differences in log-TDV and median gADC across all 10 patients fell within the RC, with a bias of -0.0251 (p=0.828) for log-TDV and -0.0047 (p=0.751) for median gADC, further confirming excellent repeatability. **Figure 5** shows automated delineations for baseline scans of two patients from the repeatability cohort (**Dataset B**).

| | **CoV (%)**<br>**[95% CIs]** | **RC**<br>**[95% CIs]** | **RC (%)**<br>**[95% CIs]** | **Sw**<br>**[95% CIs]** | **Sb**<br>**[95% CIs]** | **ICC**<br>**[95% CIs]** |
|---|---|---|---|---|---|---|
| **log-TDV** | **4.57**<br>**[3.19 - 8.01]** | **0.66**<br>**[0.46 - 1.16]** | **12.66**<br>**[8.84 - 22.2]** | **0.24**<br>**[0.17 - 0.42]** | **0.99**<br>**[0.69 - 1.73]** | **0.94**<br>**[0.81 - 0.99]** |
| Mean gADC | 4.69<br>[3.28 - 8.83] | 0.12<br>[0.09 - 0.21] | 13.0<br>[9.09 - 22.8] | 0.04<br>[0.03 - 0.08] | 0.21<br>[0.15 - 0.38] | 0.96<br>[0.85 - 0.99] |
| **Median gADC** | **3.54**<br>**[2.47 - 6.21]** | **0.08**<br>**[0.06 - 0.15]** | **9.81**<br>**[6.86 - 17.2]** | **0.03**<br>**[0.02 - 0.05]** | **0.18**<br>**[0.13 - 0.32]** | **0.97**<br>**[0.90 - 0.99]** |
| gADC variance | 33.8<br>[23.6 - 59.4] | 0.17<br>[0.12 - 0.29] | 93.8<br>[65.5 - 164] | 0.06<br>[0.04 - 0.11] | 0.14<br>[0.10 - 0.25] | 0.84<br>[0.52 - 0.96] |
| gADC skewness | 52.0<br>[36.3 - 91.2] | 1.98<br>[1.38 - 3.48] | 144<br>[101 - 253] | 0.71<br>[0.50 - 1.25] | Not calculable | Not calculable |
| gADC kurtosis | 21.8<br>[15.3 - 38.3] | 3.03<br>[2.12 - 5.32] | 60.5<br>[42.3 - 106] | 1.09<br>[0.76 - 1.92] | 2.76<br>[1.93 - 4.85] | 0.86<br>[0.57 - 0.96] |

**Table 3.** Summary of repeatability study results for automated delineations along with measurements of TDV and gADC from repeated WB-DWI baseline scans. The table shows the mode estimate for each metric, with 95% confidence intervals indicated in parentheses. The results indicated excellent repeatability for log-TDV and mean/median gADC estimates; however, a trend toward poor reproducibility was noted for higher-order gADC statistics.

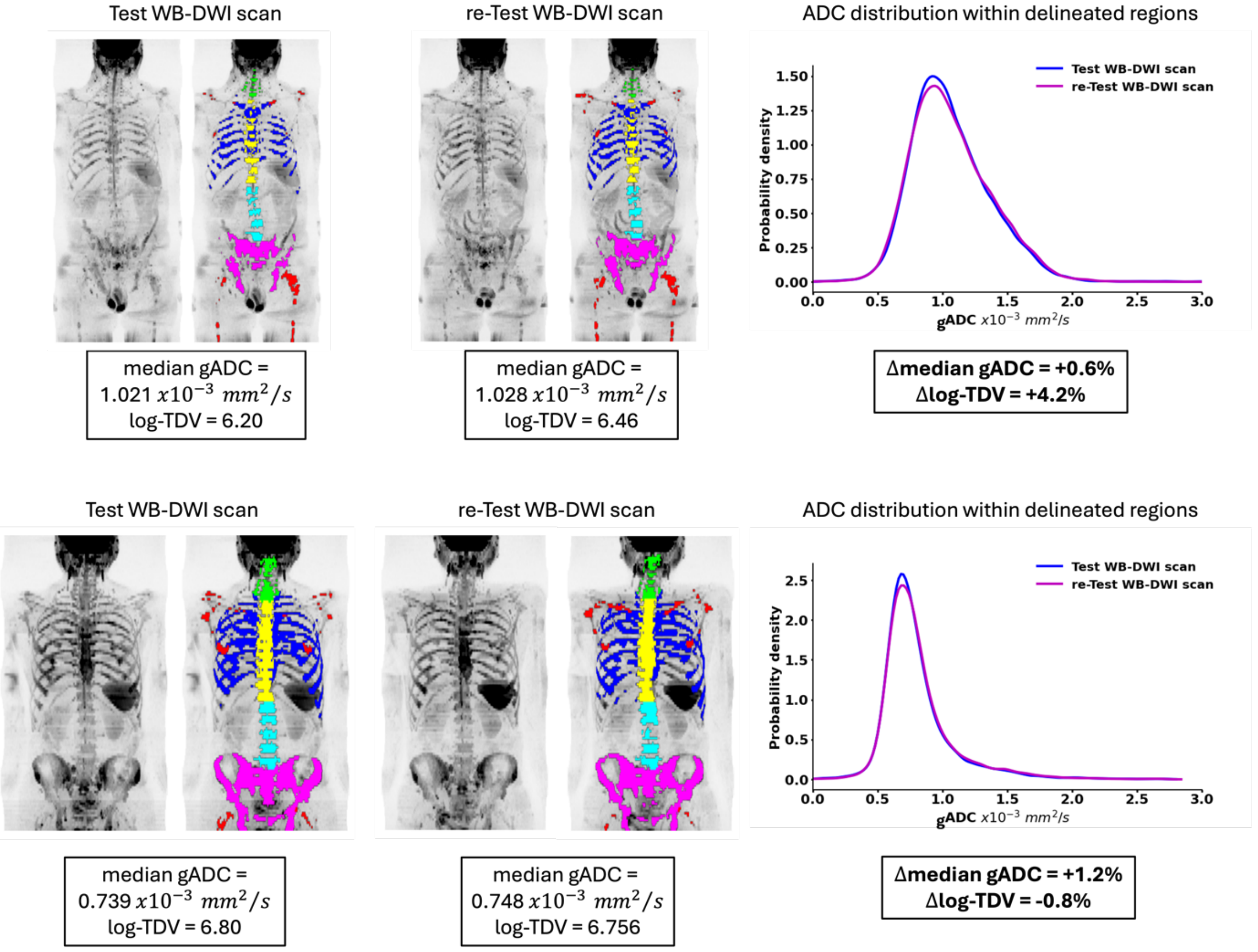

**Figure 5.** Coronal Maximum-Intensity-Projection (MIP) of signal-normalised b900 images with superimposed automated delineations for two patients from the repeatability cohort (**Dataset B**). The software delineated bony regions with high signal intensity on high b-value images and low gADC values (typically above $0.5x10^{-3}mm^2/s$) within the patient's skeleton, effectively excluding signals from normal bone marrow background, adjacent internal organs, and image background. The delineations were then transferred onto the gADC maps to derive TDV and gADC statistics. log-TDV and median gADC values derived from repeated baseline WB-DWI scans showed minor differences, demonstrating excellent repeatability. The automated delineations are color-coded by skeletal region: upper/lower limbs (red), pelvis (purple), lumbar spine (cyan), thoracic spine (yellow), cervical spine (green), and thorax (blue).

### 3.4 Accuracy in classifying response to treatment on test datasets

The optimal cutoff values derived from the bootstrapping optimisation approach were $\Delta$TDV $\leq -42.6\%$ [95% CIs: $-62$ to $-26.4$] and $\Delta$median gADC $\geq +28\%$ [95% CIs: +19 to +34] for responders (Youden's index=0.61 [95% CIs: 0.48-0.73]), and $\Delta$TDV $>+36\%$ [95% CIs: +17 to +63] for disease progression (Youden's index=0.8 [95% CIs: 0.7-0.87]). These cutoff values showed no statistically significant difference in discriminatory performance in our test **Datasets C** and **D** when compared with the pre-defined REC-based thresholds ($p<0.05$, Section 2.5.4), with an AUC of 0.82 [95% CIs: 0.73 - 0.89]).

Furthermore, we reported automated TDV and median gADC measurements from pre- and post-treatment scans for responders and patients with disease progression, as classified by the *construct reference standard* (**Figure 6**). Responders (N=43) showed a significant decrease in TDV (from 239mL [85-441] pre-treatment to 131mL [43-300] post-treatment, $p<0.005$), and a significant increase in median gADC (from 0.815x$10^{-3}$mm$^2$/s [0.709-0.912] pre-treatment to 1.01x$10^{-3}$mm$^2$/s [0.788-1.138] post-treatment, $p<0.001$). Instead, patients with disease progression (N=48) showed a significant increase in TDV (from 355mL [122-577] pre-treatment to 447mL [256-676] post-treatment, $p<0.001$). These results align with previous studies, where the same imaging biomarkers at baseline and follow-up were derived from expert-defined delineation of bone lesions in APC patients [26], [40].

Finally, we compared the software-based response classification with the *construct reference standard*. As shown in **Table 4**, our software solution achieved an accuracy of 80.4% (95% CIs: 71.6-87.4), sensitivity of 84.7% (95% CIs: 73-92.1), and specificity of 74.4% (95% CIs: 59.4-85.4) in classifying patients as *Benefit* (responder/stable) or *No Benefit* (progression) from treatment across the retrospective multi-centre cohort (**Dataset C**). In the prospective trial (**Dataset D**), the software achieved an accuracy of 81.2% (95% CIs: 56.4-94.7), with a sensitivity of 81.8% (95% CIs: 50-96.7) and a specificity of 80% (95% CIs: 34.3-99). **Figure 7** presents automated delineations and associated TDV and median gADC measurements for two test cases, one responder and one with disease progression based on the *construct reference standard*, demonstrating the value of non-invasive response imaging biomarkers. **Figure S3** in the Supplementary Material shows a test case with a low segmentation accuracy and incorrect response classification, highlighting limitations of the current software solution.

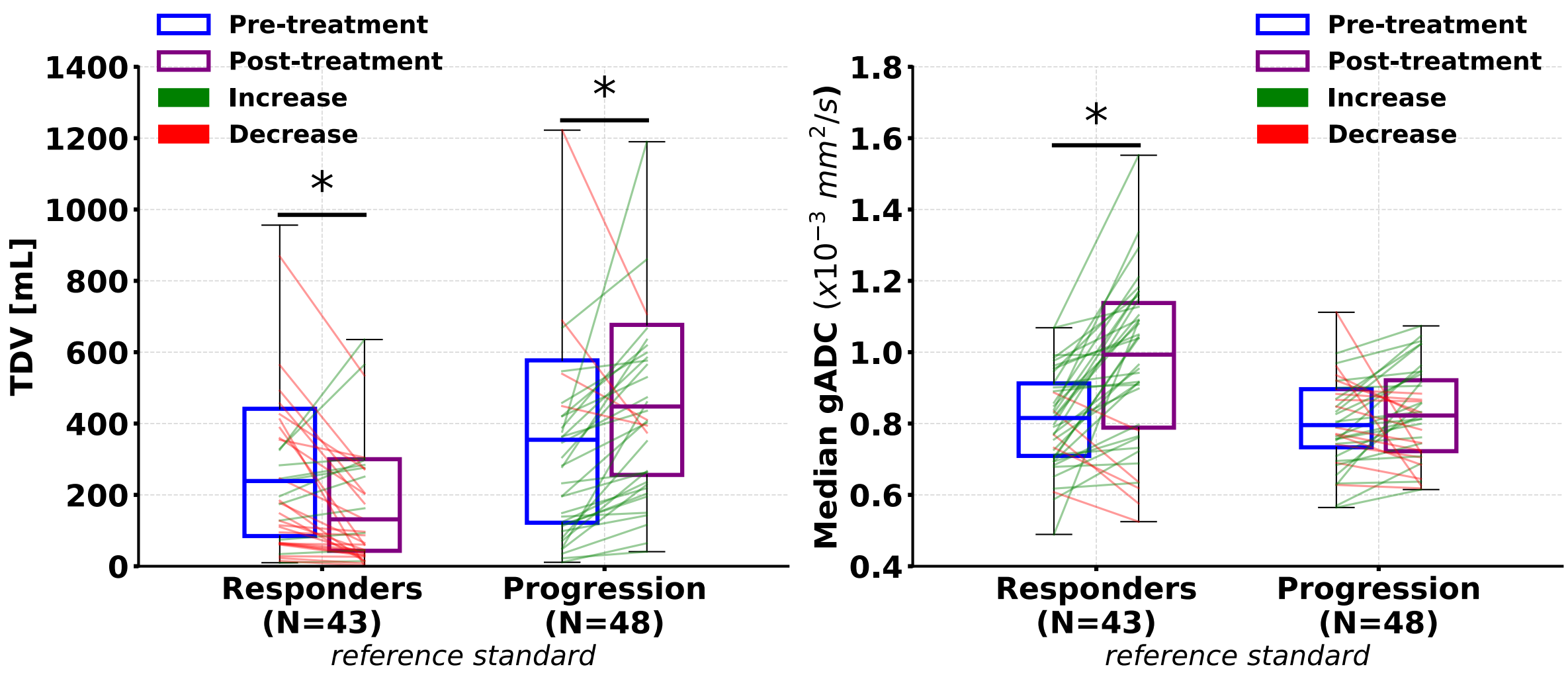

**Figure 6.** TDV and median gADC values from automated delineations of pre- and post-treatment WB-DWI scans for responders and patients with disease progression, classified according to the *reference standard* constructed by clinicians. This analysis included all patients from test cohorts (**Datasets C** and **D**). *Significant differences between pre- and post-treatment values are indicated ($p<0.05$, Wilcoxon signed-rank test: two-tailed for TDV, one-tailed for median gADC).

| | Accuracy | Sensitivity | Specificity |
|---|---|---|---|
| **Dataset (C)**<br>**(retrospective multi-center)**<br>Benefit (responders/stable) vs.<br>No Benefit (progression) | 82/102 (80.4%)<br>[95% CIs: 71.6 - 87.4] | 50/59 (84.7%)<br>[95% CIs: 73 - 92.1] | 32/43 (74.4%)<br>[95% CIs: 59.4 - 85.4] |
| **Dataset (D)**<br>**(prospective single-center)**<br>Benefit (responders/stable) vs.<br>No Benefit (progression) | 13/16 (81.2%)<br>[95% CIs: 56.4 - 94.7] | 9/11 (81.8%)<br>[95% CIs: 50 - 96.7] | 4/5 (80%)<br>[95% CIs: 34.3 - 99] |
| **ALL** | 95/118 (80.5%)<br>[95% CIs: 72.1 - 87] | 59/70 (84.3%)<br>[95% CIs: 73.7 - 91.4] | 36/48 (75%)<br>[95% CIs: 60.5 - 85.7] |

**Table 4.** Software performance in assessing treatment response in patients with APC. Accuracy, sensitivity, and specificity were calculated by comparing the software-based response assessment to the *reference standard* constructed by clinicians. Accuracy was calculated as (True Positives (TP) + True Negatives (TN)) / (Positives (P) + Negatives (N)), sensitivity TP / P, and specificity TN / N. The overall diagnostic accuracy was 80.5%, with sensitivity and specificity of 84.3 and 75%, respectively.

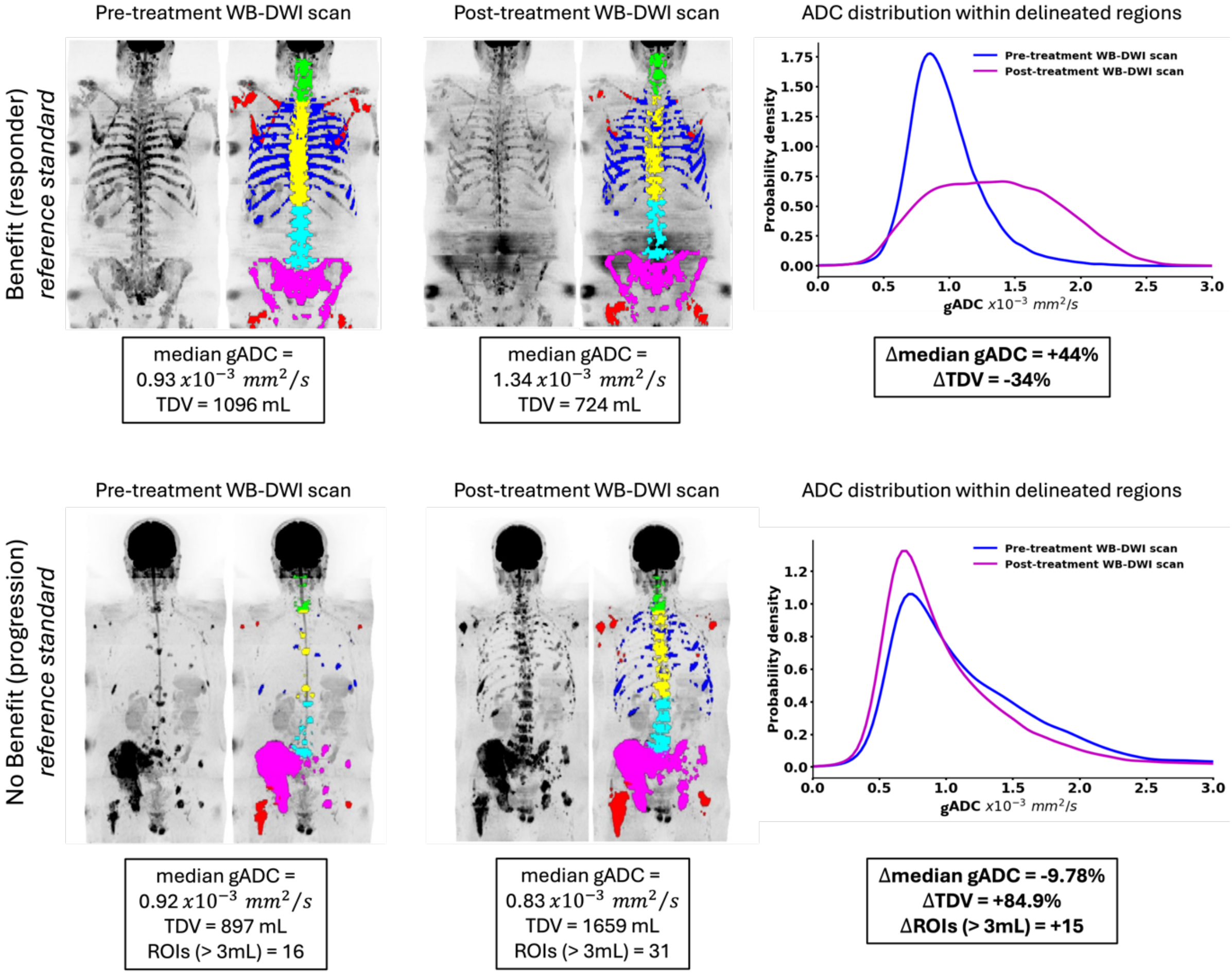

**Figure 7.** Coronal Maximum-Intensity-Projection (MIP) of signal-normalised b900 pre- and post-treatment images with superimposed automated delineations for two patients in **Dataset D**. The first patient (top row) was classified as a responder, while the second patient (bottom row) was classified as experiencing disease progression, based on the *construct reference standard*. The software delineated bony regions with high signal intensity on high b-value images and low gADC values (typically above $0.5\text{x}10^{-3}\text{mm}^2/\text{s}$) within the patient's skeleton, effectively excluding signals from normal bone marrow, adjacent internal organs, and image background, as expected. The delineations were then transferred onto the gADC map to derive TDV and gADC statistics. Using post-treatment changes in these imaging biomarkers and applying the established REC, the software-based response assessment matched the *reference standard* for both patients. For the first patient, there was a significant increase in the median gADC after treatment, exceeding the cutoff value of +25% (*Benefit*, responder). For the second patient, there was a significant increase in TDV, exceeding the cutoff value of +40% (*No Benefit*, progression), with 15 additional delineated ROIs showing volumes above 3mL between pre- and post-treatment WB-DWI scans. The automated delineations are color-coded by skeletal region: upper/lower limbs (red), pelvis (purple), lumbar spine (cyan), thoracic spine (yellow), cervical spine (green), and thorax (blue).

## 3. Discussion and Conclusions

We developed and validated a DWI signal-based AI-driven software solution to automatically delineate bony regions with restricted diffusion properties to quantify metastatic bone disease. The software automates the entire process, from image processing of pre- and post-treatment whole-body MRI studies to generating a structured report within three minutes. The report includes TDV and median gADC measurements from delineated regions, aiding response assessment following our developed REC.

Our fully automated method showed good agreement with expert-defined delineations of bone lesions detected in the pelvis and spine, with a dice score of 0.6 [0.41-0.73] and an average surface distance of 2mm. Given that 70% of bone metastases occur in these skeletal regions [3], these results are promising. Similar accuracy was observed for metastases in long bones; however, errors were more frequent in the thorax (dice score <0.35), due to coil proximity (causing enhanced signal in anterior and posterior ribs) and geometric/signal distortion from ribs on WB-DWI scans [46].

Segmentation accuracy also varied with disease extent. Patients with significant disease burden showed improved quantitative metrics, with precision of 0.6 vs. 0.1 and relative differences in median gADC of 4% vs. 18% compared with patients whose TDV was below 55 mL (corresponding to log-TDV=4 in **Figure 4**). When evaluating response to treatment, we observed that classification errors (based on qualitative assessment, as expert-defined annotations were not available for all test cases) occurred more frequently in patients with low TDV and/or a low metastatic bone lesion count. These trends suggest that our software is most reliable for monitoring disease changes in patients with moderate-to-high metastatic disease burden, when interpretation after treatment may be most challenging.

The observed performance variation across skeletal regions and with total disease burden may be explained by several confounding factors. The software may incorrectly delineate degenerative disease, nerve roots, and lymph nodes (i.e. located near the pelvic bones), which exhibit similar DWI signal to bone lesions. More prominently, false positives were observed in normal bone marrow with increased DWI signals (e.g., reactive hyperplasia, inflammatory responses, or treatment effects that increase marrow cellularity and water content). In patients presenting few lesions and low disease burden, these false positives can be significant, as they often lead to misclassification of response to treatment. This limitation arises from using a signal-based approach without considering DIXON T1w imaging, where malignant bone lesions show low fat content on calculated rFF maps [47]. However, co-registering whole-body DIXON and DWI sequences remains challenging due to differences in SNR, geometric distortions, and respiratory motion (breath-hold vs. free-breathing) [48], False negatives may also occur in lesions located at the edges of the inferior pubic rami, shoulders, and clavicles.

Despite these challenges, our method achieved state-of-the-art performance in delineating suspected bone lesions and estimating TDV and median gADC for response evaluation. Ceranka et al. [49] used an nn-UNet model with four-channel input (gADC map, T1w, b900 image, and skeleton mask) to delineate bone metastases on T1w imaging in 27 patients, reporting a patient-wise dice score of 0.33. Their method excluded the thorax and required 180 minutes to generate the final mask. Our software achieved a dice score of 0.6 on 66 test datasets with computation time of just 90 seconds per whole-body MRI study. Deeper models may still underperform in this setting due to disease heterogeneity and the limited availability of annotated datasets [50], both of which increase the risk of overfitting. Our shallow CNN, by contrast, achieved comparable accuracy while requiring far less computational resources and retaining interpretability.

Colombo et al. [27] employed a semi-automated tool for delineating bone disease in 10 APC patients, reporting an inter-observer dice score of 0.4, a 95% LoA of 10% for mean gADC, and suboptimal TDV reproducibility, with 12-23 minutes required to finalise the mask, depending on reader experience. In contrast, our method showed relative differences of log-TDV and median gADC between manual and automated methods below 5% and 9%, respectively, on test datasets.

Furthermore, the repeatability of log-TDV and median gADC was consistent with previous observer repeatability studies using semi-automated delineations of bone lesions. We reported a CoV of 4.57% and 3.54%, a RC of 12.66% and 9.81%, and an ICC of 0.94 and 0.97 for log-TDV and median gADC respectively, compared with CoV and RC values below 7% and 3%, and an ICC above 0.97 for the same WB-DWI biomarkers [51].

Finally, our software demonstrated good agreement with the *construct reference standard* for classifying patients as responders, stable, or with disease progression across 118 test datasets. The overall diagnostic accuracy was 80.5%, with sensitivity and specificity of 84.3% and 85.7%, exceeding conventional imaging modalities for assessing response to treatment in APC patients [52].

However, there are limitations to our current software:

1. Single MRI vendor testing

The software was tested on WB-DWI scans from a single MRI vendor, which may limit generalisability. Variations in MRI protocols across vendors can impact image quality, leading to decreased SNR, increased geometric distortions, and artifacts [53], potentially reducing the accuracy of automated delineations.

2. Lack of inter-site repeatability assessment

Due to practical constraints, we were unable to scan the same subject across multiple sites. Consequently, inter-site repeatability of WB-DWI biomarkers derived from automated delineations could not be evaluated. Establishing inter-site repeatability will be an essential step toward clinical translation and regulatory approval.

Although we could not test across multiple platforms and/or sites, we implemented vendor-agnostic features designed to mitigate performance loss when the pipeline is applied to other datasets, including:

- Pre-processing workflow that performs automatic resampling of b-value images and gADC map calculation, independent of scanner-specific software versions
- DWI signal intensity harmonisation and skeleton probability mapping to reduce sensitivity to protocol differences
- Optional ADC uncertainty estimation (when three b-values images are available) to flag potentially biased TDV and gADC measurements [34]
- Version-controlled source code and data, with reproducible training/testing pipelines (using Data Version Control, DVC) enabling re-training or fine-tuning of the lightweight CNN on new datasets

3. Lower specificity

The current version of our software does not make use of the relative fat fraction to further refine the classification of disease and non-disease. Disease delineation based on DWI signal intensity alone can lead to false positive results (e.g. due to T2 shine through from treated lesions, reactive cellular marrow or degenerative changes). We are currently developing a model that will include the use of relative fat fraction to improve diagnostic performance. This prototype already shown promising results for improving specificity as indicated by a ~20% decreased for relative TDV differences between manual and automated delineations, compared to the DWI-only version [54].

4. Imperfect gold-standard

This study is set up with an imperfect, although the best available gold standard regarding lesion delineation, as even expert reader bone disease delineations show significant inter-reader variability and are subject to interpretation. Future work will involve multiple expert readers with consensus review for resolving uncertainties.

5. Benchmarking against inter-reader agreement

Inter-reader dice scores for metastatic bone disease delineations by radiologists using semi-automated tools (with subsequent manual editing) were in the range of 0.29-0.76 [27], [55]. In the case of Blackledge et al. [55], inter-reader dice scores were not reported in the published manuscript; we calculated them from the same cohort, which also included four patients with advanced breast cancer. Our fully automated software achieved a dice score of 0.52 [0.34-0.65], including lesions in the cervical spine and ribcage without the need for manual editing. Moreover, voxel-wise overlap imperfections present at baseline are typically reproduced at follow-up (as demonstrated by our repeatability analysis); therefore, changes in derived TDV and gADC are likely to reflect true disease progression or response rather than artefacts, consistent with the tool's intended use for evaluating response to treatment in patients with APC from WB-DWI scans. This explains why our software maintains high response classification accuracy (80.5%) despite moderate dice scores.

6. Bone disease-only focus

The current software version does not account for disease to lymph nodes and/or soft tissue organs (e.g., liver metastases), which may co-exist within patients [56]. However, the evaluation of treatment response in lymph nodes and soft tissue is well-established and there are already software available to provide accurate and objective quantification of disease at these sites.

7. Toward multimodal response assessment

The model currently uses only two imaging channels. We plan to integrate additional imaging biomarkers and their treatment-related changes (e.g., T1w Dixon-derived fat fraction), laboratory biomarkers (e.g., Prostate-Specific Antigen, PSA, and Lactate dehydrogenase level, LDH), and measurements from soft-tissue and nodal disease to develop a multimodal model for treatment response assessment in patients with APC.

In conclusion, our automated software solution can quickly generate accurate and reproducible delineations of bony regions with restricted diffusion properties on DWI, thus providing estimations of TDV and gADC of bone disease for tracking treatment response in patients with APC. A prospective evaluation of our software is ongoing in a multi-centre national trial. The trial aims to demonstrate the value of our software output in directing treatment decisions, thereby improving patient outcomes using WB-DWI instead of conventional imaging techniques. The proposed approach is intended as a decision-support tool to assist clinicians in evaluating response to treatment in patients with APC and is meant to augment, not replace, clinical expertise.

- **Author Contributions**

  Conceptualization, A.C., M.D.B. and D.-M.K.; methodology, A.C. and M.D.B.; software, A.C. and R.H.; validation, A.C., R.H., M.D.B., N.T., and D.-M.K.; formal analysis, A.C.; resources, R.D., C.M., N.T., and D.-M.K.; data curation, A.C. and R.H.; writing-original draft preparation, A.C.; writing-review and editing, A.C., R.H., A.R., N.P., F.Z., L.D., F.C., R.D., A.D., C.M., N.T., M.D.B. and D.-M.K.; supervision, M.D.B. and D.-M.K.; project administration, A.R.; funding acquisition, M.D.B. and D.-M.K. All authors have read and agreed to the published version of the manuscript.

- **Funding**

  This project is funded by the NIHR Invention for Innovation award (Advanced computer diagnostics for whole body magnetic resonance imaging to improve management of patients with metastatic bone cancer II-LA-0216-20007).

- **Acknowledgments**

  This study represents independent research and we acknowledge support from the National Institute for Health and Care Research (NIHR) Biomedical Research Centre at The Royal Marsden NHS Foundation Trust and The Institute of Cancer Research, London, and by the Royal Marsden Cancer Charity and Cancer Research UK (CRUK) National Cancer Imaging Trials Accelerator (NCITA), and is funded by the NIHR Invention for Innovation award (Advanced computer diagnostics for whole body magnetic resonance imaging to improve management of patients with metastatic bone cancer II-LA-0216-20007). The views expressed are those of the author(s) and not necessarily those of the NIHR or the Department of Health and Social Care. This work uses data provided by patients and collected by the NHS as part of their care and support.

  The authors would also like to acknowledge Mint Medical®.

- **Informed Consent Statement**

  Patient consent was waived; study was conducted with retrospective and prospective data and researchers only had access to de-identified data. The retrospective data were analysed as part of a service evaluation (SE696), while the prospective data were collected under the WISE study, registered with ClinicalTrials.gov, ID: NCT04117594).

- **Data Availability Statement**

  Data can be shared upon request from the authors. However, this is subject to the establishment of an appropriate data-sharing agreement, given the sensitive nature of patient information involved.

- **Conflicts of interest**

A.C., R.H., and M.D.B. have submitted a patent to the UK Intellectual Property Office regarding the work described in this manuscript.

## References

[1] L. Bubendorf *et al.*, "Metastatic patterns of prostate cancer: An autopsy study of 1,589 patients," *Hum Pathol*, vol. 31, no. 5, pp. 578–583, 2000

[2] F. Macedo *et al.*, "Bone Metastases: An Overview," *Oncol Rev*, vol. 11, no. 1, 2017

[3] R. E. Coleman, "Clinical Features of Metastatic Bone Disease and Risk of Skeletal Morbidity," *Clinical Cancer Research*, vol. 12, no. 20 PART 2, pp. 6243–6250, 2006

[4] C. Parker *et al.*, "Alpha Emitter Radium-223 and Survival in Metastatic Prostate Cancer," *n engl j med*, vol. 369, pp. 213–236, 2013

[5] J. Sturge, M. P. Caley, J. Waxman, J. Sturge, M. P. Caley, and J. Waxman, "Bone metastasis in prostate cancer: emerging therapeutic strategies," *Nature Publishing Group*, vol. 8, pp. 357–368, 2011

[6] I. F. Tannock *et al.*, "Docetaxel plus Prednisone or Mitoxantrone plus Prednisone for Advanced Prostate Cancer," *N Engl J Med*, vol. 351, no. 15, pp. 1502–1512, 2004

[7] H. I. Scher *et al.*, "Trial design and objectives for castration-resistant prostate cancer: Updated recommendations from the prostate cancer clinical trials working group 3," *Journal of Clinical Oncology*, vol. 34, no. 12, pp. 1402–1418, 2016

[8] E. A. Eisenhauer *et al.*, "New response evaluation criteria in solid tumours: Revised RECIST guideline (version 1.1)," *Eur J Cancer*, vol. 45, no. 2, pp. 228–247, 2009

[9] H. L. Yang, T. Liu, X. M. Wang, Y. Xu, and S. M. Deng, "Diagnosis of bone metastases: A meta-analysis comparing 18FDG PET, CT, MRI and bone scintigraphy," *Eur Radiol*, vol. 21, no. 12, pp. 2604–2617, 2011

[10] E. Balliu *et al.*, "Comparative study of whole-body MRI and bone scintigraphy for the detection of bone metastases," *Clin Radiol*, vol. 65, no. 12, pp. 989–996, 2010

[11] G. Shen, H. Deng, S. Hu, and Z. Jia, "Comparison of choline-PET/CT, MRI, SPECT, and bone scintigraphy in the diagnosis of bone metastases in patients with prostate cancer: a meta-analysis," *Skeletal Radiol*, vol. 43, no. 11, pp. 1503–1513, 2014

[12] E. Dyrberg *et al.*, "68 Ga-PSMA-PET/CT in comparison with 18 F-fluoride-PET/CT and whole-body MRI for the detection of bone metastases in patients with prostate cancer: a prospective diagnostic accuracy study," *Eur Radiol*, vol. 29, no. 3, pp. 1221–1230, 2019

[13] A. Gafita *et al.*, "Response Evaluation Criteria in PSMA PET/CT (RECIP 1.0) in Metastatic Castration-resistant Prostate Cancer," *Radiology*, vol. 308, no. 1, 2023

[14] F. Kleiburg *et al.*, "Baseline PSMA PET/CT parameters predict overall survival and treatment response in metastatic castration-resistant prostate cancer patients," *Eur Radiol*, vol. 35, no. 7, pp. 4223–4232, 2025

[15] S. Vaz, B. Hadaschik, M. Gabriel, K. Herrmann, M. Eiber, and D. Costa, "Influence of androgen deprivation therapy on PSMA expression and PSMA-ligand PET imaging of prostate cancer patients," *Eur J Nucl Med Mol Imaging*, vol. 47, no. 1, pp. 9–15, 2020

[16] D. M. Koh and D. J. Collins, "Diffusion-weighted MRI in the body: Applications and challenges in oncology," *American Journal of Roentgenology*, vol. 188, no. 6, pp. 1622–1635, 2007

[17] A. R. Padhani *et al.*, "Rationale for Modernising Imaging in Advanced Prostate Cancer," *Eur Urol Focus*, vol. 3, no. 2–3, pp. 223–239, 2017
[18] R. Perez-lopez, D. N. Rodrigues, I. Figueiredo, J. Mateo, and D. J. Collins, "Multiparametric Magnetic Resonance Imaging of Prostate Cancer Bone Disease Correlation with Bone Biopsy Histological and Molecular Features," *Invest Radiol*, vol. 53, no. 2, 2018
[19] D. M. Koh *et al.*, "Whole-body diffusion-weighted mri: Tips, tricks, and pitfalls," *American Journal of Roentgenology*, vol. 199, no. 2, pp. 252–262, 2012
[20] A. R. Padhani, K. Van Ree, D. J. Collins, S. D'Sa, and A. Makris, "Assessing the relation between bone marrow signal intensity and apparent diffusion coefficient in diffusion-weighted MRI," *American Journal of Roentgenology*, vol. 200, no. 1, pp. 163–170, 2013
[21] R. Perez-Lopez *et al.*, "Imaging Diagnosis and Follow-up of Advanced Prostate Cancer: Clinical Perspectives and State of the Art," *Radiology*, vol. 292, no. 2, pp. 273–286, 2019
[22] A. R. Padhani, A. Makris, P. Gall, D. J. Collins, N. Tunariu, and J. S. De Bono, "Therapy monitoring of skeletal metastases with whole-body diffusion MRI," *Journal of Magnetic Resonance Imaging*, vol. 39, no. 5, pp. 1049–1078, 2014
[23] A. R. Padhani *et al.*, "METastasis Reporting and Data System for Prostate Cancer: Practical Guidelines for Acquisition, Interpretation, and Reporting of Whole-body Magnetic Resonance Imaging-based Evaluations of Multiorgan Involvement in Advanced Prostate Cancer," *Eur Urol*, vol. 71, no. 1, pp. 81–92, 2017
[24] A. R. Padhani, D. M. Koh, and D. J. Collins, "Whole-body diffusion-weighted MR imaging in cancer: Current status and research directions," *Radiology*, vol. 261, no. 3, pp. 700–718, 2011
[25] J. Kalpathy-Cramer, J. B. Freymann, J. S. Kirby, P. E. Kinahan, and A. F. W. Prior, "Quantitative Imaging Network: Data Sharing and Competitive AlgorithmValidation Leveraging the Cancer Imaging Archive," *Transl Oncol*, vol. 7, no. 1, p. 147, 2014
[26] M. D. Blackledge *et al.*, "Assessment of treatment response by total tumor volume and global apparent diffusion coefficient using diffusion-weighted MRI in patients with metastatic bone disease: A feasibility study," *PLoS One*, vol. 9, no. 4, pp. 1–8, 2014
[27] A. Colombo *et al.*, "Semi-automated segmentation of bone metastases from whole-body mri: Reproducibility of apparent diffusion coefficient measurements," *Diagnostics*, vol. 11, no. 3, 2021
[28] R. Grimm and A. R. Padhani, "Whole-body Diffusion-weighted MR Image Analysis with syngo.via Frontier MR Total Tumor Load," *MAGNETOM Flash*, vol. 68, no. 2, pp. 73–75, 2017
[29] Y. Kohada *et al.*, "Novel quantitative software for automatically excluding red bone marrow on whole-body magnetic resonance imaging in patients with metastatic prostate cancer: A pilot study," *International Journal of Urology*, vol. 30, no. 4, pp. 356–364, 2023
[30] R. Donners *et al.*, "Inter- and Intra-Patient Repeatability of Radiomic Features from Multiparametric Whole-Body MRI in Patients with Metastatic Prostate Cancer," *Cancers (Basel)*, vol. 16, no. 9, p. 1647, 2024
[31] J. Mateo *et al.*, "Olaparib in patients with metastatic castration-resistant prostate cancer with DNA repair gene aberrations (TOPARP-B): a multicentre, open-label, randomised, phase 2 trial," *Lancet Oncology*, vol. 21, no. 1, pp. 162–174, 2020

[32] J. M. Winfield, M. D. Blackledge, N. Tunariu, D. M. Koh, and C. Messiou, "Whole-body MRI: a practical guide for imaging patients with malignant bone disease," *Clin Radiol*, vol. 76, no. 10, pp. 715–727, 2021

[33] M. D. Blackledge *et al.*, "Computed Diffusion-weighted MR Imaging May Improve Tumor Detection," *Radiology*, vol. 261, no. 2, pp. 573–81, 2011

[34] M. D. Blackledge *et al.*, "Noise-Corrected, Exponentially Weighted, Diffusion-Weighted MRI (niceDWI) Improves Image Signal Uniformity in Whole-Body Imaging of Metastatic Prostate Cancer," *Front Oncol*, vol. 10, pp. 1–12, 2020

[35] A. Candito *et al.*, "Deep learning assisted atlas-based delineation of the skeleton from Whole-Body Diffusion Weighted MRI in patients with malignant bone disease," *Biomed Signal Process Control*, vol. 92, p. 106099, 2024

[36] A. Candito *et al.*, "A weakly-supervised deep learning model for fast localisation and delineation of the skeleton, internal organs, and spinal canal on Whole-Body Diffusion-Weighted MRI (WB-DWI)," 2025

[37] A. Candito *et al.*, "Deep Learning for Delineation of the Spinal Canal in Whole-Body Diffusion-Weighted Imaging: Normalising Inter- and Intra-Patient Intensity Signal in Multi-Centre Datasets," *Bioengineering*, vol. 11, no. 2, p. 130, 2024

[38] J. M. Winfield *et al.*, "Extracranial Soft-Tissue Tumors: Repeatability of Apparent Diffusion Coefficient Estimates from Diffusion-weighted MR Imaging," *Radiology*, vol. 284, no. 1, 2017

[39] J. M. Bland and D. G. Altman, "Statistical methods for assessing agreement between two methods of clinical measurement," *Int J Nurs Stud*, vol. 47, no. 8, pp. 931–936, 2010

[40] R. Perez-Lopez *et al.*, "Diffusion-weighted imaging as a treatment response biomarker for evaluating bone metastases in prostate cancer: A pilot study," *Radiology*, vol. 283, no. 1, pp. 168–177, 2017

[41] C. Parker *et al.*, "Radium-223 in metastatic castration-resistant prostate cancer: whole-body diffusion-weighted magnetic resonance imaging scanning to assess response," *Cancer Spectrum*, vol. 7, no. 6, 2023

[42] A. Rutjes, J. B. Reitsma, A. Coomarasamy, K. S. Khan, and P. Bossuyt, "Evaluation of diagnostic tests when there is no gold standard. A review of methods," *Health Technol Assess (Rockv)*, vol. 11, no. 50, 2007

[43] D. L. Raunig *et al.*, "Quantitative imaging biomarkers: A review of statistical methods for technical performance assessment," *Statistical Methods in Medical Research*, vol. 24, no. 1. SAGE Publications Ltd, pp. 27–67, 2015

[44] C. Messiou, D. J. Collins, S. Giles, J. S. De Bono, D. Bianchini, and N. M. De Souza, "Assessing response in bone metastases in prostate cancer with diffusion weighted MRI," *Eur Radiol*, vol. 21, no. 10, pp. 2169–2177, 2011

[45] I. Lavdas *et al.*, "Apparent diffusion coefficient of normal abdominal organs and bone marrow from whole-body DWI at 1.5 T: The effect of sex and age," *American Journal of Roentgenology*, vol. 205, no. 2, pp. 242–250, 2015

[46] S. Keaveney *et al.*, "Image quality in whole-body MRI using the MY-RADS protocol in a prospective multi-centre multiple myeloma study," *Insights Imaging*, vol. 14, no. 1, pp. 1–14, 2023

[47] R. Donners, M. Blackledge, N. Tunariu, C. Messiou, E. M. Merkle, and D.-M. Koh, "Quantitative Whole-Body Diffusion - Weighted MR Imaging," *Magn Reson Imaging Clin N Am*, vol. 26, no. 4, pp. 479–495, 2018

[48] M. Rata *et al.*, “Implementation of Whole-Body MRI (MY-RADS) within the OPTIMUM/MUKnine multi-centre clinical trial for patients with myeloma,” *Insights Imaging*, vol. 13, no. 1, pp. 1–16, 2022

[49] J. Ceranka, F. Lecouvet, J. De Mey, and J. Vandemeulebroucke, “Computer-aided detection of focal bone metastases from whole-body multi-modal MRI,” no. March 2020, p. 27, 2020

[50] Z. Huang *et al.*, “Revisiting model scaling with a U-net benchmark for 3D medical image segmentation,” *Sci Rep*, vol. 15, no. 29795, 2025

[51] R. Donners *et al.*, “Repeatability of quantitative individual lesion and total disease multiparametric whole-body MRI measurements in prostate cancer bone metastases,” *British Journal of Radiology*, vol. 96, no. 1151, 2023

[52] N. Tunariu *et al.*, “METRADS-P vs. RECIST/PCWG criteria to detect disease progression in metastatic castration-resistant prostate cancer (mCRPC),” *Journal of Clinical Oncology*, vol. 42, no. 16, 2024

[53] A. Barnes *et al.*, “Guidelines & recommendations: UK quantitative WB-DWI technical workgroup: Consensus meeting recommendations on optimisation, quality control, processing and analysis of quantitative whole-body diffusion-weighted imaging for cancer,” *British Journal of Radiology*, vol. 91, no. 1081, pp. 1–12, 2018

[54] A. Candito *et al.*, “Developing a multi-channel deep-learning model for automatically quantifying malignant bone disease from Multiparametric Whole-Body MRI,” *The International Society for Magnetic Resonance in Medicine (ISMRM)* 2025

[55] M. D. Blackledge *et al.*, “Inter- and Intra-Observer Repeatability of Quantitative Whole-Body, Diffusion-Weighted Imaging (WBDWI) in Metastatic Bone Disease,” *PLoS One*, vol. 11, no. 4, pp. 1–12, 2016

[56] N. Tunariu *et al.*, “What’s New for Clinical Whole-body MRI (WB-MRI) in the 21st Century,” *Br J Radiol*, vol. 93, pp. 1–13, 2020